\documentclass{assets/template}
\PassOptionsToPackage{numbers, compress, sort}{natbib}

\usepackage[utf8]{inputenc} 
\usepackage[T1]{fontenc}    
\usepackage{hyperref}       
\usepackage{url}            
\usepackage{booktabs}       
\usepackage{amsfonts}       
\usepackage{nicefrac}       
\usepackage{microtype}      
\usepackage{xcolor}         
\usepackage{natbib}
\usepackage{enumitem}
\usepackage{tcolorbox}
\usepackage{pifont}
\usepackage{listings}
\usepackage{textcomp}

\usepackage{booktabs}
\usepackage{colortbl}
\usepackage{xcolor}
\usepackage{ulem} 
\usepackage{graphicx} 
\definecolor{rowgray}{HTML}{E8F0FE}
\definecolor{headbg}{HTML}{D6E4F0}
\definecolor{avgbg}{HTML}{BDD7EE}

\usepackage{threeparttable}
\usepackage{graphicx}
\usepackage{wrapfig}
\usepackage{float}
\usepackage{bbm}
\usepackage{amsmath}
\usepackage{amssymb}
\usepackage{mathtools}
\usepackage{amsthm}
\usepackage{amsthm}
\usepackage{cleveref}
\usepackage{hyperref}
\usepackage{multirow}
\usepackage{subcaption}
\usepackage{tikz}

\newcommand*\circled[1]{%
  \tikz[baseline=(char.base)]{
    \node[shape=circle,draw,inner sep=1pt] (char) {#1};
  }
}

\usepackage[ruled,vlined]{algorithm2e}
\usepackage{algorithmic}  
\usepackage{etoc}

\title{Memory Grafting: Scaling Language Model Pre-training via Offline Conditional Memory}

\vspace{-18pt}
\author{%
Runxi Cheng\textsuperscript{\rm 1}
\quad 
Yuchen Guan\textsuperscript{\rm 1}
\quad
Yongxian Wei\textsuperscript{\rm 1}
\quad 
Qianpu Sun\textsuperscript{\rm 1}
\quad 
Qixiu Li\textsuperscript{\rm 1}
\\
\textbf{
Sinan Du\textsuperscript{\rm 1}
\quad
Feng Xiong\textsuperscript{\rm 1}
\quad 
Chun Yuan\textsuperscript{\rm 1}$^{\dagger}$
\quad 
Yan Lu\textsuperscript{\rm 2}$^{\dagger}$
\quad 
Yeyun Gong\textsuperscript{\rm 2}$^{\dagger}$
}
\\
\textsuperscript{\rm 1}Tsinghua University, \quad
\textsuperscript{\rm 2}Microsoft Reasearch Asia
\quad $^\dagger$ Corresponding Authors\\
{\tt\small crx23@mails.tsinghua.edu.cn; \quad yegong@microsoft.com}
}

\begin{document}

\begin{abstract}
Scaling conditional memory offers a promising way to increase language-model
capacity, but existing methods such as Engram learn large memory tables from
scratch during pre-training, making memory scaling expensive and sometimes
ineffective. We propose \textit{Memory Grafting}, a conditional memory scaling method
that utilize frozen hidden states from a grafting model as conditional
$n$-gram memory. Given frequent local $n$-grams, we run the grafting model
offline, store final-token hidden representations as memory values, and let
the recipient model retrieve them through exact longest-match suffix lookup.
Retrieved memories are adapted by lightweight projections and gates, while a
hash-based Engram fallback preserves coverage for unmatched contexts. Since
the grafting model is only run offline and exact lookup has expected
$\mathcal{O}(1)$ complexity with respect to memory-bank size, Memory Grafting
expands external latent capacity with limited training and inference overhead.
Experiments under matched recipient architectures and pre-training budgets show
that Memory Grafting improves over both MoE and vanilla Engram baselines. In the
2.8B-scale setting, it improves the average benchmark score from
$51.95$ for MoE and $52.43$ for vanilla Engram to $53.86$. In the
0.92B--scale setting, all grafting-model variants improve over
the baselines, with Qwen3.5-35B-A3B giving the strongest gains.  These results suggest that pretrained models can serve as reusable
constructors of external latent memory, providing a practical step toward
scaling future language models beyond trainable parameters alone.
\end{abstract}

\maketitle

\section{Introduction}
\label{sec:introduction}

The rapid progress of large language models~\citep{achiam2023gpt,team2025kimi,team2023gemini,liu2024deepseek,glm2024chatglm} over the past few years
has been driven in large part by joint scaling of model parameters
and pre-training
tokens~\citep{kaplan2020scaling,hoffmann2022training}. As frontier
models have grown to hundreds of billions to trillions of parameters
trained on trillions of tokens, the engineering cost of each
pre-training run has risen substantially. In response, recent work
has increasingly focused on architectural sparsity, where only a
fraction of a model's total capacity participates in any individual
forward pass. Mixture-of-Experts (MoE)
architectures~\citep{shazeer2017outrageously,fedus2022switch,
dai2024deepseekmoe} are a representative instance and underlie a number
of recent open-weight systems. Another direction~\citep{lample2019large,ultramem,engram} scales
capacity through sparse memory lookup rather than sparse computation,
attaching a large external table that is queried at each token. We refer to this line of work as
conditional memory.

\noindent Conditional memory offers a scalable way to expand model capacity while adding
only limited inference overhead. In particular, Engram~\cite{engram} uses
deterministic hash-based addressing to support $\mathcal{O}(1)$
retrieval per token, and reports that offloading a
$100$B-parameter memory only costs around $2\%$
end-to-end inference throughput on a $4$B-dense backbone. However, the
training side is less favorable. The memory table is learned from
scratch during pre-training, so scaling it increases the parameters need to be optimized. Empirically, training efficiency began to decline once the
memory module accounts for more than $70\%$ of total
parameters. Thus,
scaling Engram still demands substantial training-time computation.

\begin{figure*}[t!]
	\centering
	\includegraphics[width=0.9\linewidth]{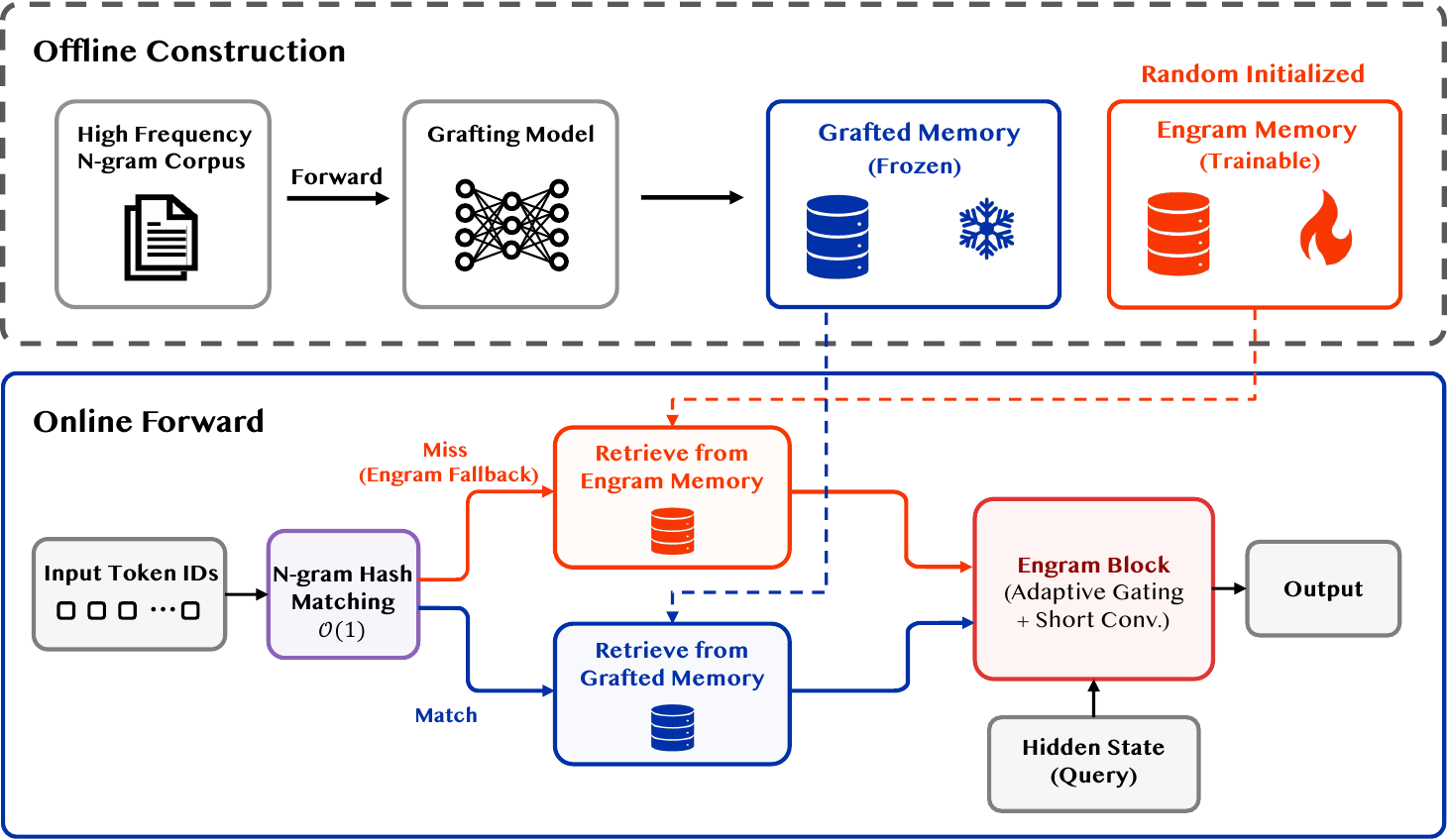}
	\caption{Computation pipeline for Memory Grafting}
        \label{Computation pipeline}
    \vspace{-16pt}
\end{figure*}

\noindent To solve this problem, we revisit what Engram memory stores~\cite{shannon1948mathematical,kneser1995improved,engram}: an
$n$-gram-indexed table of static embeddings. Such entries naturally
correspond to recurring multi-token constructions -- named
entities, formulaic phrases, library calls, and mathematical
notation~\citep{liu2024infini}
-- whose meaning is largely fixed by surface form. A sufficiently
pre-trained model has already learned strong representations for
these patterns. If these
encodings can be used as conditional memory, the memory capacity can
scale without increasing the recipient's training cost, while the
$\mathcal{O}(1)$ lookup that makes conditional memory efficient at
inference remains intact.

\noindent Motivated by this observation, we propose Memory Grafting, which
constructs conditional memory from a grafting model rather than
learning the memory table entirely from scratch. We run the grafting
model once offline over frequent local $n$-grams and store the
resulting hidden states as frozen memory values. During recipient
training and inference, the model retrieves these values through
longest-match exact lookup, falls back to hash-based Engram memory
on misses, and injects the selected memory through lightweight
projection and gating. Since the grafting-model computation
is paid once offline lightly, the input sequence length is unchanged and
the per-token lookup remains $\mathcal{O}(1)$, the conditional
memory can be scaled while keeping both training and inference efficient.  Our contributions are summarized as follows:
\begin{itemize}
    \item We introduce \textbf{Memory Grafting}, a representation-level
    knowledge-transfer paradigm for scaling conditional memory. It
    stores frozen grafting-model hidden states in an exact
    $n$-gram-indexed table, retains a hash-based Engram fallback for
    unmatched contexts, and grafts retrieved memories through a
    gated residual update. The grafting-model cost is paid once
    offline lightly, and per-token lookup
    remains $\mathcal{O}(1)$.
    \item We evaluate Memory Grafting under matched recipient
    architectures and pre-training budgets. The results show gains
    over MoE and vanilla Engram baselines, increasing exact-memory
    coverage as the table grows, practical inference efficiency, and
    probing evidence that the recipient actually uses the grafted
    memory.
\end{itemize}

\section{Related Work}
\subsection{Conditional Memory and Modular Sparse Scaling}
Efficiently scaling model capacity is crucial for advancing the performance of large language models. Sparse scaling and conditional memory represent two prominent directions in this pursuit. Mixture-of-Experts  architectures \citep{shazeer2017outrageously, fedus2022switch, dai2024deepseekmoe}, which activate only a fraction of the total capacity during any single forward pass, have become a widely adopted sparse paradigm in recent open-weight systems. Parallel to sparse computation, another line of research focuses on scaling capacity through sparse memory lookups rather than computation, a mechanism collectively termed conditional memory~\citep{gemma_3n_2025,gemma_4,engram}. For instance, Product-Key Memory \citep{lample2019large} replaces standard feed-forward layers with a massive key-value store, enabling sparse nearest-neighbor search via Cartesian product sets. Building on this, UltraMem \citep{ultramem} resolves sparse retrieval bottlenecks and scaling inefficiencies through tucker decomposed query-key retrieval and implicit value expansion. STEM~\cite{sadhukhan2026stem} replaces the FFN up-projection in Transformer layers with token-indexed embedding modules, reducing per-token computation while increasing parameter capacity through input-dependent embedding lookup. MoLE~\cite{jie2025mixture} improves MoE inference efficiency by training experts as FFNs and re-parameterizing them into input-id-based lookup tables before inference, thereby reducing expert computation and offloading overhead while preserving MoE-like capacity. More recently, Engram \citep{engram} learns an $n$-gram vocabulary as its memory module, employing hash-based addressing to achieve $\mathcal{O}(1)$ sparse retrieval during the forward pass.  Such conditional memory offers a clear
inference-side advantage, but scaling it still requires substantial
computation when the memory itself is learned from scratch during pre-training.


\subsection{Retrieval-Augmented Generation and Knowledge Distillation}

Beyond conditional memory, another line of work improves language models by
leveraging external memory sources in different ways. Retrieval-Augmented
Generation~\citep{lewis2020retrieval} queries external non-parametric memories
or documents during training or inference, with pioneering works such as
REALM~\citep{guu2020retrieval} casting both pre-training and fine-tuning as an
end-to-end retrieve-then-predict process, and recent
work~\citep{islam2024open,lin2024rella} further refining retrieval reasoning and
extraction quality. In contrast, Knowledge Distillation transfers the
parametric memory of a larger teacher into a recipient model, improving its
capability without explicit retrieval at inference
time~\citep{peng2025pre,goyal2025distilled}.

\noindent Although effective, RAG and KD introduce different efficiency bottlenecks. RAG
relies on retrieval and longer augmented contexts at inference, shifting
substantial overhead to deployment and increasing latency~\cite{zhang2024accelerating,jiang2023chameleon,gao2023retrieval}. KD instead
internalizes external knowledge into the student's parameters, but logits-based
distillation during from-scratch pre-training requires running the teacher
across the entire corpus and consuming large soft-label distributions, incurring
substantial compute and memory cost. In contrast, Memory Grafting
needs only a lightweight one-time offline pass to build the grafted memory
table, preserving high efficiency in both training and inference and showing
strong potential for model scaling.

\section{Method}
\label{sec:method}

\subsection{Preliminaries on Engram}
\label{sec:prelim-engram}

Engram~\cite{engram} is a conditional memory module for language models. Unlike MoE,
which sparsely activates computation experts, Engram sparsely retrieves
memory indexed by local token patterns. Given a token sequence
$\mathbf{X}=(x_1,\ldots,x_T)$ and hidden states $\mathbf{H}^{(\ell)}\in\mathbb{R}^{T\times D}$
at layer $\ell$, Engram performs two steps: hashed $N$-gram retrieval and
context-aware fusion.

\noindent For each position $t$, Engram forms suffix $n$-grams from the local history.
Following the original design, token ids may first be mapped through a
vocabulary compression function $P: \mathcal{V}\rightarrow\mathcal{V}'$, so
that textually equivalent tokens share a canonical id. The compressed suffix
$n$-gram is:
\begin{equation}
    \mathbf{g}_{t,n} = \big(P(x_{t-n+1}),\ldots,P(x_t)\big),
    \qquad n\in\{2,\ldots,N\}.
\end{equation}
Because enumerating all possible $N$-grams is infeasible, Engram uses
multi-head hashing. For each order $n$ and hash head $k$, a deterministic
hash function $\phi_{n,k}$ maps $\mathbf{g}_{t,n}$ to an embedding-table row:
\begin{equation}
    z_{t,n,k}=\phi_{n,k}(\mathbf{g}_{t,n}),
    \qquad
    \mathbf{e}_{t,n,k}=\mathbf{E}_{n,k}[z_{t,n,k}].
\end{equation}
The retrieved embeddings across all $n$-gram orders and hash heads are
concatenated into a static memory vector:
\begin{equation}
    \mathbf{e}_t = \mathop{\Vert}_{n=2}^{N}\mathop{\Vert}_{k=1}^{K} \mathbf{e}_{t,n,k}.
\end{equation}
Since $\mathbf{e}_t$ is retrieved only from the local token pattern, it is not by
itself aware of the global context. Engram therefore uses the current hidden
state as a query to gate the retrieved memory. For a branch $c$ of the
residual stream, let $\mathbf{h}_{t}^{(c)}\in\mathbb{R}^{D}$ be the hidden state. The
memory vector is projected into key and value vectors:
\begin{equation}
    \mathbf{k}_t^{(c)} = \mathbf{W}_K^{(c)} \mathbf{e}_t,
    \qquad
    \mathbf{v}_t = \mathbf{W}_V \mathbf{e}_t .
\end{equation}
The gate is computed by comparing the normalized key with the normalized
query~\cite{zhang2019root,elfwing2018sigmoid,engram}:
\begin{equation}
    \alpha_t^{(c)}  = \sigma \left (
    \frac{\left\langle \mathrm{RMSNorm}(\mathbf{k}_t^{(c)}),
    \mathrm{RMSNorm}(\mathbf{h}_t^{(c)})\right\rangle}{\sqrt{D}} \right )
\end{equation}
The gated value is then
processed by a lightweight depthwise causal convolution:
\begin{equation}
    \mathbf{u}_t^{(c)}=\alpha_t^{(c)}\mathbf{v}_t,
    \qquad
    \Delta\mathbf{h}_t^{(c)} = \mathbf{u}_t^{(c)} + \mathrm{ShortConv}(\mathbf{U}^{(c)})_t.
\end{equation}

\begin{algorithm}[t!]
\caption{Memory Grafting with Engram Fallback}
\label{alg:memory_grafting}
\begin{algorithmic}[1]
\STATE\textbf{Input:} Hidden states $\mathbf{H}\in\mathbb{R}^{B\times T\times C\times D}$, tokens $\mathbf{X}$, frozen memory $(\mathcal{M},\mathbf{W}_{\mathrm{mem}})$, trainable Engram table $\mathbf{W}_{\mathrm{hash}}$, projections $\mathbf{W}^{\mathrm{mem}}_{k,v},\mathbf{W}^{\mathrm{eng}}_{k,v}$
\STATE\textbf{Output:} Grafted hidden states $\widetilde{\mathbf{H}}$

\STATE $\mathbf{M}_{\mathrm{mem}}\leftarrow\mathbf{0}$, $\mathbf{m}\leftarrow\mathbf{0}$
\hfill {// Initialize memory features and hit mask}
\FOR{each token position $(b,t)$}
  \STATE $j\leftarrow\texttt{ExactLookup}(\mathbf{X}_{b,\leq t},\mathcal{M})$
  \hfill {// Select the largest matched n-gram}
  \IF{$j$ exists}
    \STATE $\mathbf{M}_{\mathrm{mem}}[b,t]\leftarrow\mathbf{W}_{\mathrm{mem}}[j]$, $\mathbf{m}[b,t]\leftarrow1$
  \ENDIF
\ENDFOR

\FOR{each token position $(b,t)$}
  \IF{$\mathbf{m}[b,t]=1$}
    \STATE $\mathbf{e}_{b,t}\leftarrow\mathbf{M}_{\mathrm{mem}}[b,t]$, \quad $(\mathbf{K}_{b,t},\mathbf{V}_{b,t})\leftarrow(\mathbf{W}^{\mathrm{mem}}_{k}\mathbf{e}_{b,t},\mathbf{W}^{\mathrm{mem}}_{v}\mathbf{e}_{b,t})$
    \hfill {// Memory grafting path}
  \ELSE
    \STATE $\mathbf{z}_{b,t}\leftarrow\texttt{NgramHashMapping}(\mathbf{X}_{b,\leq t})$
    \hfill {// Generate deterministic hash ids}
    \STATE $\mathbf{E}_{b,t}\leftarrow\texttt{MultiHeadEmbedding}(\mathbf{z}_{b,t})$
    \hfill {// Lookup trainable Engram table}
    \STATE $\mathbf{e}_{b,t}\leftarrow\texttt{Concat}(\mathbf{E}_{b,t})$
    \STATE $(\mathbf{K}_{b,t},\mathbf{V}_{b,t})\leftarrow(\mathbf{W}^{\mathrm{eng}}_{k}\mathbf{e}_{b,t},\mathbf{W}^{\mathrm{eng}}_{v}\mathbf{e}_{b,t})$
    \hfill {// Engram fallback path}
  \ENDIF
\ENDFOR

\STATE $\mathbf{Q}\leftarrow\mathrm{RMSNorm}_{q}(\mathbf{H})$, \quad $\bar{\mathbf{K}}\leftarrow\mathrm{RMSNorm}_{k}(\mathbf{K})$
\STATE $\boldsymbol{\alpha}\leftarrow\sigma(\langle\bar{\mathbf{K}},\mathbf{Q}\rangle/\sqrt{D})$
\hfill {// Query-key gate}
\STATE $\mathbf{U}\leftarrow\boldsymbol{\alpha}\odot\mathbf{V}$, \quad $\Delta\mathbf{H}\leftarrow\mathbf{U}+\texttt{ShortConv}(\mathbf{U})$
\STATE $\widetilde{\mathbf{H}}\leftarrow\mathbf{H}+\Delta\mathbf{H}$
\STATE \textbf{return} $\widetilde{\mathbf{H}}$
\end{algorithmic}
\end{algorithm}
    \vspace{-12pt}

\subsection{Memory Grafting}
\label{sec:memory-grafting}
Original Engram learns a hash-indexed memory table from scratch. Memory
Grafting instead constructs an external memory bank from latent
representations of a grafting model and grafts the retrieved features into a
recipient model. This changes the role of conditional memory from purely
trainable storage to reusable representation-level knowledge transfer.
\paragraph{Frozen Latent Memory Construction.}
We first build a set of frequent $n$-gram keys
$\mathcal{M}=\{\mathbf{m}_i\}_{i=1}^{M}$, where each $\mathbf{m}_i$ is a token-id tuple such as
a 2-, 3-, or 4-gram. For each key $\mathbf{m}_i$, we run a grafting model $F_G$ offline
and extract a hidden representation of its original text from a selected grafting layer. In our
implementation, we use the hidden state at the final token of the $n$-gram as
the memory embedding. This produces a frozen table
\begin{equation}
    \mathbf{W}_{\mathrm{mem}}[i] = F_G(\texttt{text}(\mathbf{m}_i))_{\mathrm{last}}^{(r)}
    \in \mathbb{R}^{D_{\mathrm{mem}}},
\end{equation}
where $r$ is the grafting layer used for extraction. The mapping from $\mathbf{m}_i$ to
row index $i$ is stored as an exact integer lookup structure. The table
$\mathbf{W}_{\mathrm{mem}}$ is not updated during recipient-model training;
only the recipient-side projection and gating parameters are learned.

\paragraph{Exact Retrieval with Longest-Match Priority.}
At a grafting layer of the recipient model, let
$\mathbf{H}\in\mathbb{R}^{B\times T\times C\times D}$ denote the hidden
states, where $B$ is the batch size, $T$ is the sequence length, $C$ is the
number of residual branches, and $D$ is the branch dimension. For each
position $(b,t)$, Memory Grafting queries the frozen memory using suffix
$n$-grams ending at $x_{b,t}$. If multiple keys match the same position, we
keep the longest matched $n$-gram, e.g., a 4-gram match overrides a 3-gram match. This yields a memory feature $\mathbf{E}_{\mathrm{mem},b,t}$ and a binary
hit mask $m_{b,t}\in\{0,1\}$:
\begin{equation}
    (j_{b,t}, m_{b,t}) = \texttt{ExactLookup}(\mathbf{X}_{b,\leq t},\mathcal{M}),
    \qquad
    \mathbf{E}_{\mathrm{mem},b,t} = \mathbf{W}_{\mathrm{mem}}[j_{b,t}]
    \quad \text{if } m_{b,t}=1.
\end{equation}
The exact lookup is independent of the total memory size in expectation under
hash-map or trie-style indexing, and costs only a constant number of suffix
queries per token because the maximum $n$-gram order is fixed.

\paragraph{Memory Grafting with Engram Fallback.}
The frozen memory covers only the selected frequent $n$-grams. To preserve
full coverage over arbitrary inputs, we retain the original hash-based Engram
path as a fallback. For every position, we compute a trainable hash Engram
feature:
\begin{equation}
    \mathbf{E}_{\mathrm{hash},b,t} = \texttt{Concat}\big(
    \texttt{MultiHeadEmbedding}(\texttt{NgramHashMapping}(\mathbf{X}_{b,\leq t}))
    \big).
\end{equation}
Memory-hit positions use the frozen grafted memory, whereas memory-miss
positions use the trainable Engram fallback. We use separate projections for
the two sources:
\begin{align}
    \mathbf{K}_{\mathrm{mem},b,t}^{c} &= \mathbf{W}_{k}^{\mathrm{mem},c} \mathbf{E}_{\mathrm{mem},b,t},
    &
    \mathbf{V}_{\mathrm{mem},b,t} &= \mathbf{W}_{v}^{\mathrm{mem}} \mathbf{E}_{\mathrm{mem},b,t}, \\
    \mathbf{K}_{\mathrm{eng},b,t}^{c} &= \mathbf{W}_{k}^{\mathrm{eng},c} \mathbf{E}_{\mathrm{hash},b,t},
    &
    \mathbf{V}_{\mathrm{eng},b,t} &= \mathbf{W}_{v}^{\mathrm{eng}} \mathbf{E}_{\mathrm{hash},b,t}.
\end{align}
The branch-specific key and shared value used by the gate are selected by the
hit mask:
\begin{equation}
    \mathbf{K}_{b,t}^{c} = m_{b,t} \mathbf{K}_{\mathrm{mem},b,t}^{c}
    +(1-m_{b,t}) \mathbf{K}_{\mathrm{eng},b,t}^{c},
    \qquad
    \mathbf{V}_{b,t} = m_{b,t} \mathbf{V}_{\mathrm{mem},b,t}
    +(1-m_{b,t}) \mathbf{V}_{\mathrm{eng},b,t}.
\end{equation}

\paragraph{Gated Grafting.}
The selected memory value is grafted into the decoder using the same
context-aware gating principle as Engram. For each branch $c$, we compute
\begin{equation}
    \alpha_{b,t}^{c}= \sigma \left(
    \frac{\left\langle \bar{\mathbf{K}}_{b,t}^{c},\mathbf{Q}_{b,t}^{c}\right\rangle}{\sqrt{D}} \right ),\;\text{where}
    \;     \mathbf{Q}_{b,t}^{c}=\mathrm{RMSNorm}_{q}(\mathbf{H}_{b,t}^{c}),
    \;
    \bar{\mathbf{K}}_{b,t}^{c}=\mathrm{RMSNorm}_{k}(\mathbf{K}_{b,t}^{c})
\end{equation}
The output can be calculated as:
\begin{equation}
    \mathbf{U}_{b,t}^{c}=\alpha_{b,t}^{c}\mathbf{V}_{b,t},
    \qquad
    \Delta\mathbf{H} = \mathbf{U} + \texttt{ShortConv}(\mathbf{U}),
    \qquad
    \widetilde{\mathbf{H}}=\mathbf{H}+\Delta\mathbf{H}.
\end{equation}

\section{Experiments}
\label{sec:experiments}
\begin{table}[t!]
\centering
\caption{Benchmark performance (\%) in the 2.8B-trainable, 100B-token setting.}
\label{tab:benchmark}
\setlength{\tabcolsep}{10pt}
\renewcommand{\arraystretch}{1.25}
\small
\resizebox{0.9\textwidth}{!}{
\begin{tabular}{@{}lc*{3}{c}}
\toprule
\textbf{Benchmark} \textit{\scriptsize(Metric)} & \textbf{\# Shots} & \textbf{MoE Baseline} & \textbf{Vanilla Engram} & \textbf{Memory Grafting} \\
\midrule
\# Trainable Params                           & & 2.8B          & 2.8B          & 2.8B \\
\# Activated \textit{\scriptsize(w/o token embed)} & & 0.55B          & 0.55B          & 0.55B \\
\# Trained Tokens                         & & 100B           & 100B           & 100B \\
\# Experts \textit{\scriptsize(shared + routed, top-$k$)} & & 1 + 64 (top-4) & 1 + 48 (top-4) & 1 + 47 (top-4) \\
\midrule
ARC-Challenge    & 0-shot  & 36.12 & 36.09 & \cellcolor{rowgray}\textbf{37.03} \\
ARC-Easy         & 0-shot  & 72.39 & 71.80 & \cellcolor{rowgray}\textbf{73.40} \\
BoolQ            & 0-shot  & 57.92 & 60.24 & \cellcolor{rowgray}\textbf{62.54} \\
Social IQA       & 0-shot  & 41.61 & 41.76 & \cellcolor{rowgray}\textbf{42.94} \\
RACE             & 0-shot  & 34.74 & 34.55 & \cellcolor{rowgray}\textbf{35.98} \\
LAMBADA          & 5-shot  & 42.98 & 45.22 & \cellcolor{rowgray}\textbf{48.19} \\
WinoGrande       & 5-shot  & 58.98 & 59.66 & \cellcolor{rowgray}\textbf{60.93} \\
PIQA             & 5-shot  & 75.41 & 75.63 & \cellcolor{rowgray}\textbf{76.17} \\
HellaSwag        & 10-shot & 47.39 & 46.94 & \cellcolor{rowgray}\textbf{47.54} \\
\midrule
\textbf{Average} & & 51.95 & 52.43 & \cellcolor{rowgray}\textbf{53.86} \\
\bottomrule
\end{tabular}
}

\end{table}

\begin{table}[t!]
\centering
\caption{Result in the 0.92B-trainable, 50B-token setting
with different grafting models. MG-$X$ denotes Memory Grafting using model $X$
as the grafting model: MG-GLM refers to GLM-4.7-Flash, MG-Deepseek refers to
DeepSeek-V2-Lite, and MG-Qwen3.5 refers to Qwen3.5-35B-A3B.}
\label{tab:benchmark-full}
\setlength{\tabcolsep}{3pt}
\renewcommand{\arraystretch}{1.2}
\small
\resizebox{0.9\textwidth}{!}{
\begin{tabular}{@{}lc*{5}{c}}
\toprule
\textbf{Benchmark} \textit{\scriptsize(Metric)} & \textbf{\# Shots} & \textbf{MoE Baseline} & \textbf{Vanilla Engram} & \textbf{MG-GLM} & \textbf{MG-Deepseek} & \textbf{MG-Qwen3.5} \\
\midrule
\# Trainable Params                           & & 0.92B          & 0.92B          & 0.92B          & 0.92B          & 0.92B \\
\# Activated \textit{\scriptsize(w/o token embed)} & & 0.29B          & 0.29B          & 0.29B          & 0.29B          & 0.29B \\
\# Trained Tokens                         & & 50B            & 50B            & 50B            & 50B            & 50B \\
\# Experts \textit{\scriptsize(shared + routed, top-$k$)} & & 1 + 64 (top-4) & 1 + 48 (top-4) & 1 + 46 (top-4) & 1 + 46 (top-4) & 1 + 46 (top-4) \\
\midrule
ARC-Challenge    & 0-shot  & 27.56          & 28.92          & 28.07          & 28.92          & \cellcolor{rowgray}\textbf{29.35} \\
ARC-Easy         & 0-shot  & 64.31          & 64.31          & 62.96          & \textbf{64.39} & \cellcolor{rowgray}64.06 \\
BoolQ            & 0-shot  & 55.93          & 46.97          & 53.03          & 56.64          & \cellcolor{rowgray}\textbf{58.10} \\
Social IQA       & 0-shot  & 39.36          & 40.28          & 40.12          & 40.48          & \cellcolor{rowgray}\textbf{40.84} \\
RACE             & 0-shot  & \textbf{31.77} & 31.39          & 30.05          & 30.62          & \cellcolor{rowgray}31.58 \\
LAMBADA          & 5-shot  & 31.88          & 32.27          & \textbf{34.37} & 32.52          & \cellcolor{rowgray}33.73 \\
WinoGrande       & 5-shot  & 51.07          & 51.85          & 53.51          & 53.51          & \cellcolor{rowgray}\textbf{55.01} \\
PIQA             & 5-shot  & 70.67          & 70.95          & 71.87          & \textbf{71.93} & \cellcolor{rowgray}71.44 \\
HellaSwag        & 10-shot & 38.02          & 38.34          & 38.48          & \textbf{38.80} & \cellcolor{rowgray}38.74 \\
\midrule
\textbf{Average} & & 45.62 & 45.03 & 45.83 & 46.42 & \cellcolor{rowgray}\textbf{46.98} \\
\bottomrule
\end{tabular}
}

\end{table}
\subsection{Experimental Setup}
\label{exp setup}
\textbf{Model Architectures.} We implement two recipient model scales~\cite{vaswani2017attention}, with
total parameters around $0.9$B and $2.8$B. For both scales, the MoE layers
contain $64$ routed experts, and following DeepSeekMoE~\citep{dai2024deepseekmoe},
we use one shared expert in each MoE layer. For the Engram module, we follow the
best parameter allocation reported in the Engram paper and set the parameter
ratio between Engram memory and MoE computation to $1{:}3$. Since Memory
Grafting introduces additional lightweight projection parameters for adapting
the frozen memory features to the recipient hidden space, we reduce the
corresponding number of experts to keep the fair comparison. Across all
experiments, we ensure that the number of trainable parameters follows
Memory Grafting $\le$  vanilla Engram $\le$  MoE baseline. 

\noindent \textbf{Data \& Tokenizer.} We trained the 925M parameter model on a 50B-token subset of the 
Nemotron-CC dataset~\citep{su2025nemotron} and the $2.81$B parameter model on a 100B-token subset of the Nemotron-CC dataset. All text was tokenized with the LLaMA-3-8B tokenizer~\citep{grattafiori2024llama}. 
\begin{figure*}[t!]
	\centering
	\includegraphics[width=1.0\linewidth]{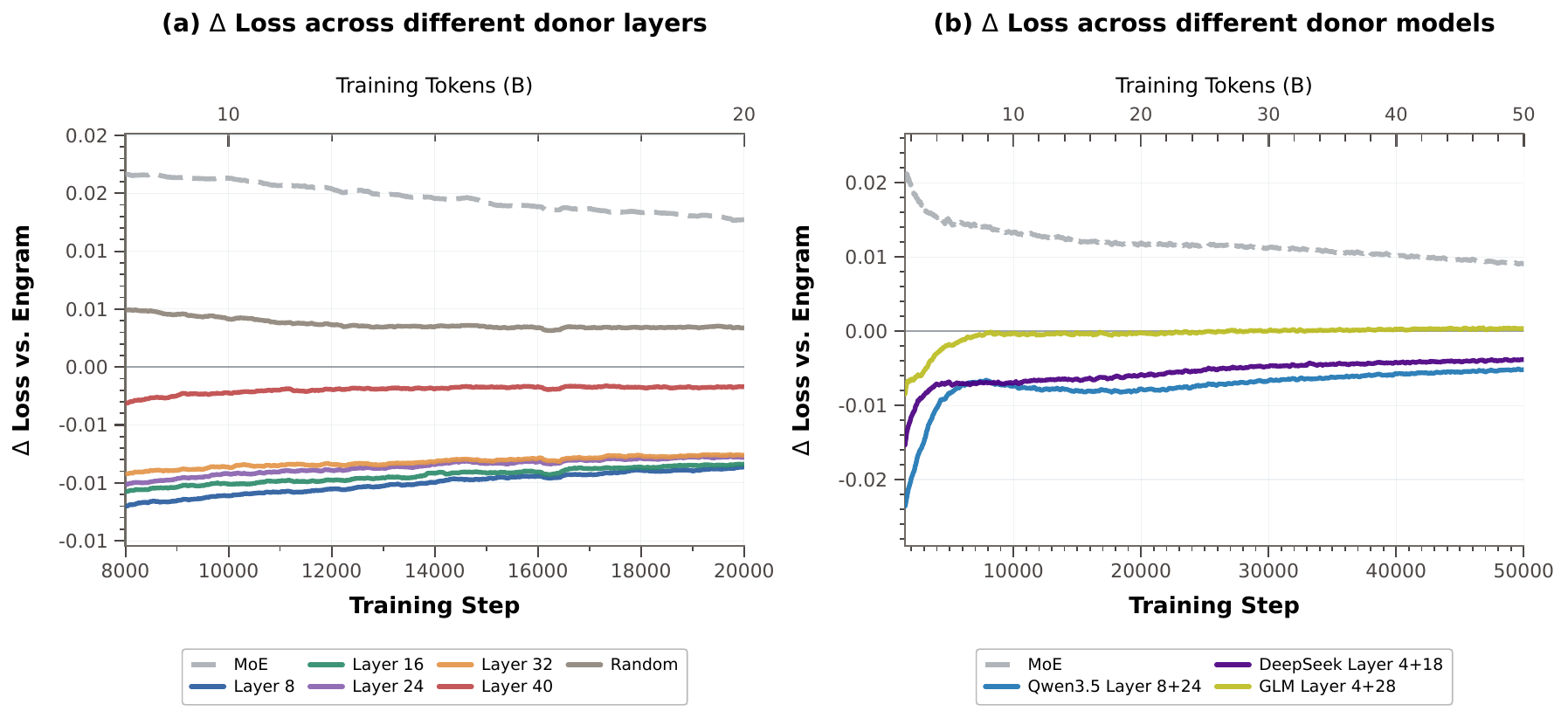}
\caption{Loss differences relative to vanilla Engram when varying the
	grafted memory source.  (a) Comparison across different Qwen3.5-35B-A3B source layers (b) Comparison across different grafting models.}

        \label{compare_activation}
\end{figure*}

\noindent \textbf{Grafting Setting.}  We use Qwen3.5-35B-A3B~\cite{qwen3.5}, DeepSeek-V2-Lite~\cite{deepseekv2}, and
GLM-4.7-Flash~\cite{zeng2025glm} as grafting models. For each $n$-gram order, we select the top
$1$M most frequent $n$-grams and compute their frozen memory representations
offline, yielding $3$M entries per grafted layer. Each entry stores a
hidden-state vector of the grafting model in \texttt{bfloat16}; for
Qwen3.5-35B-A3B , this corresponds to roughly $12$\,GB of frozen memory per layer. To cover different representation levels of the grafting model, we extract hidden states from an
early-to-middle layer and a deeper layer, approximately corresponding to
$10\%$ and $60\%$ of the model depth. In particular, we use Layers 8, 24 for
Qwen3.5-35B-A3B, Layers 4, 18 for DeepSeek-V2-Lite, and Layers 4, 28 for
GLM-4.7-Flash.

\noindent \textbf{Hyper-Parameters.} The hyper-parameters are selected based on the common practice for dense language models. We replace all FFN layers with MoE layer in the transformer. Please refer \Cref{exp_detail} for detailed training hyper-parameters.

\noindent \textbf{Benchmarks.} We use the lm-evaluation-harness~\citep{eval-harness} for evaluation. The benchmarks used include ARC-C~\citep{arc-c},ARC-E\citep{arc-c}, BoolQ~\citep{boolq}, HellaSwag~\citep{hellaswag}, LAMBADA~\citep{lambada},  PIQA~\citep{piqa}, RACE~\citep{lai2017race}, WinoGrande~\citep{sakaguchi2021winogrande}, .

\subsection{Main Results}
\label{sec:main-results}

We report two main sets of experimental results. Table~\ref{tab:benchmark} uses a 2.8B
total~/~0.55B activated MoE backbone trained for 100B tokens
and compares the MoE baseline, vanilla Engram, and Memory Grafting with a
Qwen3.5-35B-A3B grafting model. Memory Grafting is best on all nine
benchmarks: the average improves from $51.95$ for MoE and $52.43$ for vanilla
Engram to $53.86$. The largest gains appear on LAMBADA,
from $42.98$ to $48.19$; BoolQ, from $57.92$ to $62.54$; and WinoGrande,
from $58.98$ to $60.93$. Since the grafting memory is frozen and produced
offline, and the recipient keeps matched trainable parameters and activated
parameters, the result demonstrate the effectiveness of memory grafting.

\noindent  Table~\ref{tab:detailed_arch_1B} fixes a 0.92B trainable~/~0.29B
activated recipient trained for 50B tokens and varies only
the grafting model. All grafting variants improve average accuracy over the
baselines, but the gains are uneven: MG-GLM gives only a small improvement,
from $45.62$ for MoE to $45.83$, whereas MG-Deepseek and MG-Qwen3.5 reach
$46.42$ and $46.98$, respectively. We also observe that vanilla Engram
underperforms the MoE baseline in this 1B-scale setting, with $45.03$ versus
$45.62$. These may caused by the limited Engram embedding table in 0.9B model, which lead to more hash collisions. This suggests that Memory Grafting benefits from external
latent memory, but the quality of the grafting model's $n$-gram
representations matters. Appendix~\ref{app:grafting-model-memory-geometry}
analyzes the memory geometry and shows that GLM-4.7-Flash memory is much more
anisotropic and less discriminative, which helps explain why its downstream
gain is limited.

\subsection{Further Analysis}
\subsubsection{Impact of the Grafting-Model Source Layer}
\label{sec:further-analysis-grafting-source-layer}
Beyond the choice of grafting model, we find that the grafting-model layer used
to construct the frozen memory is also important. In this analysis, we keep the
recipient model, training data, memory keys, and recipient-side grafting module
fixed, and vary only the source layer from which the grafting-model $n$-gram
embeddings are extracted. The loss trend shows that memory built from an
intermediate grafting-model layer, especially around Layer 6 in the compared
setting, gives the strongest benefit. When the memory embeddings are extracted
from deeper grafting-model layers, the loss can instead increase.

\noindent This result suggests that Memory Grafting requires a compatible match between
the representation level of the external memory and the recipient model's own
hidden states. Earlier or middle-layer grafting-model embeddings are more
likely to encode local lexical and phrase-level $n$-gram information that a
small recipient model can align with. Deeper grafting-model embeddings may be
more abstract and more tightly coupled to the internal computation of the large
grafting model. For a smaller recipient, these high-level $n$-gram
representations can be harder to interpret through a lightweight projection and
gate, making the injected signal less useful or even harmful. Therefore, the
benefit of Memory Grafting depends not only on whether external memory is
available, but also on choosing a grafting-model source layer whose
representations the recipient can effectively understand.

\begin{figure*}[t!]
	\centering
	\includegraphics[width=\linewidth]{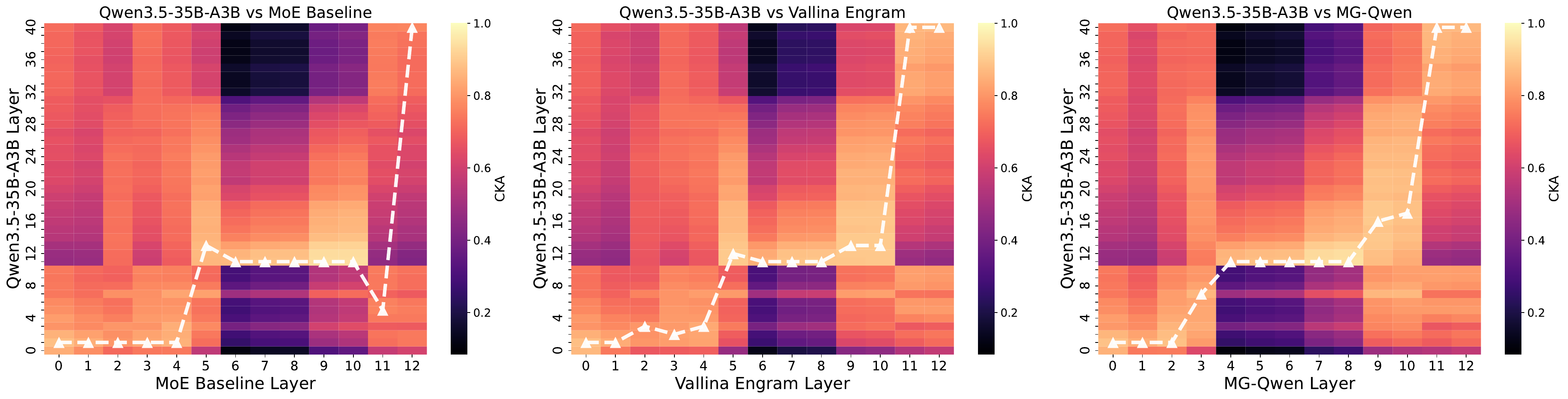}
	\caption{CKA comparison between Qwen3.5-35B-A3B and the MoE baseline,
	vanilla Engram, and Memory Grafting model. The white trajectory marks the
	best-matching recipient layer for each Qwen3.5-35B-A3B layer.}

	\label{fig:cka-memory-grafting}
\end{figure*}
\subsubsection{CKA Analysis of Memory Grafting}
\label{sec:further-analysis-cka-memory-grafting}

We use centered kernel alignment (CKA)~\cite{davari2022reliability} to analyze how Memory Grafting
changes the recipient representation space, as shown in
Figure~\ref{fig:cka-memory-grafting}. The heatmaps compare
Qwen3.5-35B-A3B layers with the layers of the MoE baseline, vanilla Engram,
and Memory Grafting model, and the white trajectory marks the best-matching
recipient layer for each grafting-model layer. Compared with the two
baselines, Memory Grafting shows a clearer alignment with deeper
Qwen3.5-35B-A3B layers at shallower recipient layers. In other words, after
memory grafting, the recipient begins to exhibit functions similar to the deep
Qwen representations earlier in its own layer stack, whereas the MoE baseline
and vanilla Engram show weaker or later alignment. This indicates that the
grafted memory leaves a measurable signature in the recipient hidden states, which supports the intended mechanism of Memory Grafting. Specifically, the recipient-side projection and gate allow the recipient model to
effectively incorporate feature patterns injected from the grafting model,
thereby aligning its internal representations with the grafted memory.

\subsubsection{Ablation of Grafting Mechanism}
\begin{wrapfigure}[14]{r}{0.48\textwidth}
    \centering
    \includegraphics[width=\linewidth]{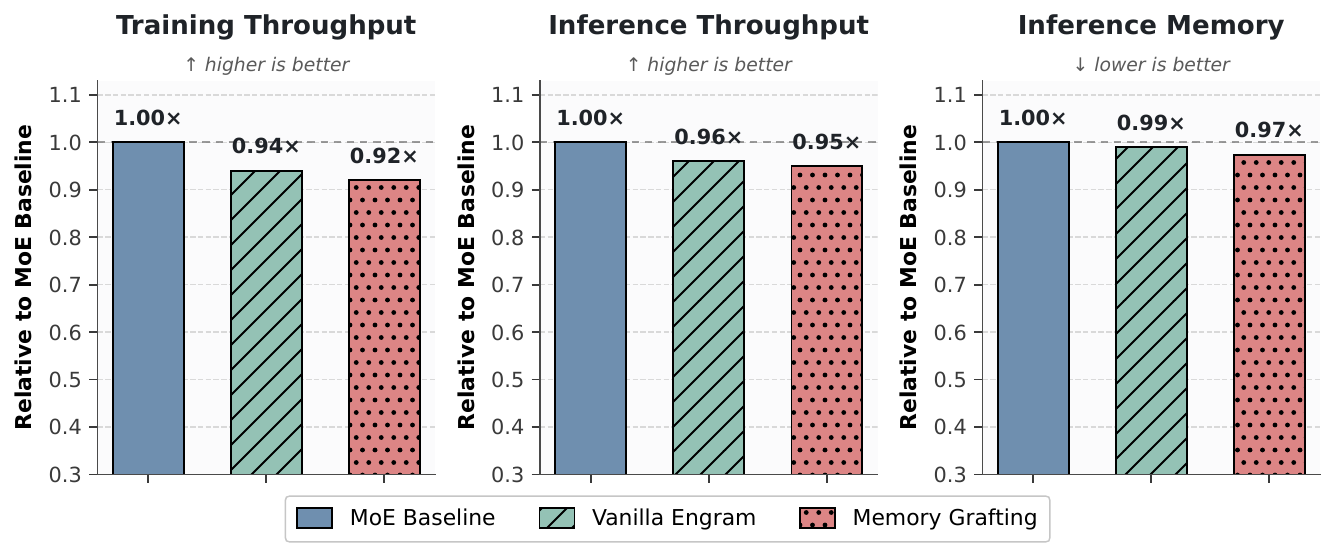}
    \caption{Throughput and memory usage comparison. The model of Memory Grafting has 0.92B trainable parameters with approximately 12B frozen grafting memory offloaded.}
    \label{fig:speed}
    \vspace{-10pt}
\end{wrapfigure}
\begin{figure*}[t!]
	\centering
	\includegraphics[width=\linewidth]{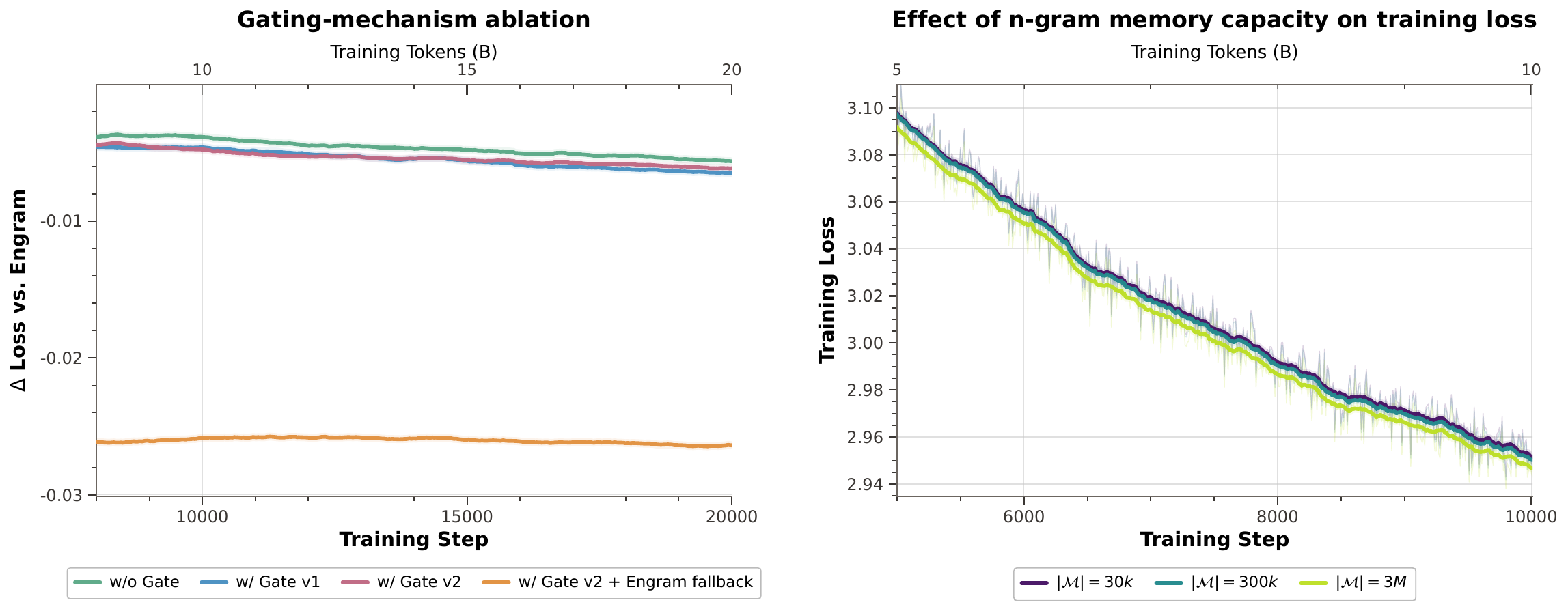}
	\caption{Ablation of the Memory Grafting design and the grafted $n$-gram
	memory capacity.}

        \label{fig:memory-capacity-scaling}
    
\end{figure*}

\label{sec:further-analysis-grafting-mechanism-ablation}
We compare four memory grafting mechanisms in
Figure~\ref{fig:memory-capacity-scaling}: directly grafting multiple
matched $n$-gram embeddings through attention, adding a gate after this
attention aggregation, grafting only the longest matched $n$-gram through a
gate, and further adding the Engram fallback to the gated longest-match
mechanism. The two attention-based variants improve over vanilla Engram, but
their gains are limited. Selecting only the longest matched
$n$-gram and grafting it through a gate is more effective because the selected
memory is more specific and easier to regulate.

\noindent The best result is obtained by further adding the Engram fallback to this gated
longest-match grafting mechanism. This variant consistently achieves the
largest loss reduction relative to vanilla Engram. Since the frozen grafting
memory is built only for high-frequency $n$-grams, it cannot cover all tokens
through exact matching. Using only this memory may therefore introduce a
hit-token bias during training: tokens with matched high-frequency contexts
receive grafting signals, while unmatched or low-frequency contexts receive no
external memory. The Engram fallback mitigates this coverage bias by providing a
trainable memory path for missed tokens and by retaining additional
low-frequency $n$-gram memories. Therefore, the final Memory Grafting mechanism
combines adaptive gating, efficient longest-match grafting, and engram fallback
mechanism.

\subsubsection{Impact of Grafted Memory Capacity}
\label{sec:further-analysis-memory-capacity-scaling}
For exploring the scaling potential of memory grafting, We compare the training loss across different grafted memory size. As shown in the right of Figure~\ref{fig:memory-capacity-scaling},
larger memory capacity consistently leads to lower training loss. Increasing
the memory size from $30$k to $300$k and further to $3$M improves the loss
curve throughout training, indicating a clear scaling trend: grafting more
memory provides more useful exact-match contexts and improves the recipient
model more strongly. The result demonstrate the benefit of Memory Grafting can be scaled with the size of
the grafted memory.
\subsubsection{Efficiency Analysis}
\label{sec:further-analysis-efficiency}
We report the efficiency comparison in
Figure~\ref{fig:speed}. Memory Grafting keeps the
runtime cost close to the MoE baseline, reaching $0.92\times$ training
throughput and $0.95\times$ inference throughput, while using $0.97\times$
inference memory. Although Memory Grafting introduces an external memory bank
that is several times larger than the recipient model, the resulting slowdown is
limited. The main reason is that exact $n$-gram retrieval is implemented as a
hash lookup, whose expected complexity is  $\mathcal{O}(1)$ with respect to the memory-bank
size. The expensive grafting-model computation is also performed offline, so the
online path only adds lookup, projection, gating, and the memory update. The
offline construction cost is also modest in our setting: extracting a $3$M
$n$-gram memory from Qwen3.5-35B-A3B requires only about $3$ A100 GPU-hours.
Therefore, the result suggests that Memory Grafting can scale to substantially
larger memory banks while adding only modest offline construction cost and
runtime overhead.


\section{Conclusion}
\label{sec:conclusion}

In this paper, we introduced \textit{Memory Grafting}, a conditional memory scaling
method that turns frozen grafting-model hidden states into conditional
$n$-gram memory for a smaller recipient model. The key idea is to move the
expensive grafting-model computation offline: the recipient performs exact
longest-match retrieval over the frozen memory bank, projects and gates the
retrieved latent representation, and falls back to trainable Engram memory when
no exact match is available. Under matched pre-training settings, this design
improves over both MoE and vanilla Engram baselines, and the gains are stronger
when the grafted memory comes from more discriminative grafting models such as
Qwen3.5-35B-A3B and DeepSeek-V2-Lite. The analysis further shows that the
retrieved memories are actively used by the recipient and that increasing
memory capacity improves training loss. Since exact memory retrieval has
expected $\mathcal{O}(1)$ complexity with respect to memory-bank size, Memory Grafting
mainly expands external latent capacity rather than per-token activated
computation, enabling efficient training and inference. We therefore view it
as a practical step toward scaling future language models through reusable
external latent memory beyond trainable parameters alone.

\clearpage
\bibliography{ref}
\bibliographystyle{ieee_fullname}

\clearpage
\appendix

\part*{}
\addcontentsline{toc}{part}{Appendix}

\begin{center}
    \LARGE \bfseries {Appendix}
\end{center}

\vspace{0.5em}

\setcounter{tocdepth}{3}
\etocsetnexttocdepth{subsubsection}
\localtableofcontents

\clearpage

\section{Experiment Details}
\label{exp_detail}
\subsection{ Model Architecture and Hyper Parameters}
\begin{table}[h!]
\centering

\caption{Architecture and training hyper-parameters for the 0.92B-class models.}

\resizebox{0.95\linewidth}{!}{%

\begin{tabular}{@{}l ccc@{}}
\toprule
 & \textbf{MoE-1B} & \textbf{Engram-1B} & \textbf{MG-1B} \\
\midrule
Total Params       & 0.92B & 0.92B & 0.92B \\
Active Params      & 0.29B & 0.29B & 0.29B \\
Total Tokens       & 50B & 50B & 50B \\
\midrule
Layers             & 12 & 12 & 12 \\
Dimension          & 768 & 768 & 768 \\
MoE Inter. Size    & 384 & 384 & 384 \\
Routed Experts     & 64 & 48 & 46 \\
Active Experts     & 4 & 4 & 4 \\
Shared Experts     & 1 & 1 & 1 \\
\midrule
Heads              & 16 & 16 & 16 \\
Sequence Length    & 2048 & 2048 & 2048 \\
Vocab Size         & 128256 & 128256 & 128256 \\
Training Steps     & 50000 & 50000 & 50000 \\
Optimizer          & AdamW & AdamW & AdamW \\
Base LR            & 5e-4 & 5e-4 & 5e-4 \\
LR Scheduler       & Cosine w/ warmup & Cosine w/ warmup & Cosine w/ warmup \\
Weight Decay       & 0.1 & 0.1 & 0.1 \\
\midrule
Engram Dim $d_\text{mem}$         & - & 384            & 384 \\
Engram $N$-gram                   & - & [2,3]          & [2,3]\,\textsuperscript{$\dagger$} \\
Engram Num Head                   & - & 8              & 8 \\
Engram Vocab Size                 & - & 112865         & 118265 \\
Engram Layer (target)             & - & [1, 6]         & [1, 6] \\
Engram ShortConv Kernel           & - & 4              & 4 \\
Engram Weight Decay               & - & 0.1            & 0.1 \\
Engram Optimizer                  & - & AdamW~\cite{loshchilov2017decoupled} & AdamW \\
\midrule
Memory Grafting Model                       & - & - & Qwen3.5-35B-A3B \\
Memory Grafting Source Layers               & - & - & [8, 24] \\
Memory Grafting Recipient Layers            & - & - & [1, 12] \\
Memory Grafting Entries per Layer           & - & - & 3{,}000{,}000 \\
Memory Grafting Hidden Dim                  & - & - & 2048 \\
Memory Grafting Dtype                       & - & - & bfloat16 \\
Memory Grafting Size per Layer              & - & - & $\approx$12\,GB \\
Memory Grafting Total Frozen Memory         & - & - & $\approx$24\,GB \\
Memory Grafting Trainable                   & - & - & False (frozen) \\
Memory Grafting Retrieval                   & - & - & Exact longest-match (2/3/4-gram) \\
Memory Grafting Fallback                    & - & - & Hash Engram (2/3-gram) \\
\bottomrule
\end{tabular}}

\label{tab:detailed_arch_1B}
\end{table}

\clearpage

\begin{table}[h!]
\centering
\caption{Architecture and training hyper-parameters for the 2.8B-class models.}
\resizebox{0.95\linewidth}{!}{%
\begin{tabular}{@{}l ccc@{}}
\toprule
 & \textbf{MoE-2.8B} & \textbf{Engram-2.8B} & \textbf{MG-2.8B} \\
\midrule
Total Params       & 2.8B & 2.8B & 2.8B \\
Active Params      & 0.55B & 0.55B & 0.55B \\
Total Tokens       & 100B & 100B & 100B \\
\midrule
Layers             & 24 & 24 & 24 \\
Dimension          & 1024 & 1024 & 1024 \\
MoE Inter. Size    & 512 & 512 & 512 \\
Routed Experts     & 64 & 48 & 47 \\
Active Experts     & 4 & 4 & 4 \\
Shared Experts     & 1 & 1 & 1 \\
\midrule
Heads              & 16 & 16 & 16 \\
RoPE $\theta$      & 10000 & 10000 & 10000 \\
Sequence Length    & 2048 & 2048 & 2048 \\
Vocab Size         & 128256 & 128256 & 128256 \\
Training Steps     & 50000 & 50000 & 50000 \\
Optimizer          & AdamW & AdamW & AdamW \\
Base LR            & 5e-4 & 5e-4 & 5e-4 \\
LR Scheduler       & Cosine w/ warmup & Cosine w/ warmup & Cosine w/ warmup \\
Weight Decay       & 0.1 & 0.1 & 0.1 \\
\midrule
Engram Dim $d_\text{mem}$         & - & 512            & 512  \\
Engram $N$-gram                   & - & [2,3]          & [2,3] \\
Engram Num Head                   & - & 8              & 8 \\
Engram Vocab Size                 & - & 292680         & 292680 \\
Engram Layer (target)             & - & [1, 12]        & [1, 12] \\
Engram ShortConv Kernel           & - & 4              & 4 \\
Engram Weight Decay               & - & 0.1            & 0.1 \\
Engram Optimizer                  & - & AdamW          & AdamW \\
\midrule
Memory Grafting Model                       & - & - & Qwen3.5-35B-A3B \\
Memory Grafting Source Layers               & - & - & [8, 24] \\
Memory Grafting Recipient Layers            & - & - & [1, 12] \\
Memory Grafting Entries per Layer           & - & - & 3{,}000{,}000 \\
Memory Grafting Hidden Dim                  & - & - & 2048 \\
Memory Grafting Dtype                       & - & - & bfloat16 \\
Memory Grafting Size per Layer              & - & - & $\approx$12\,GB ($\approx$6.1B params) \\
Memory Grafting Total Frozen Memory         & - & - & $\approx$24\,GB ($\approx$12.3B params) \\
Memory Grafting Trainable                   & - & - & False (frozen) \\
Memory Grafting Retrieval                   & - & - & Exact longest-match (2/3/4-gram) \\
Memory Grafting Fallback                    & - & - & Hash Engram (2/3-gram) \\
\bottomrule
\end{tabular}}

\label{tab:detailed_arch_3B}
\end{table}
\clearpage
\subsection{Construction of Grafting-Model Memory}
\label{app:grafting-model-memory-construction}

We construct the frozen grafting-model memory from frequent local $n$-grams
before recipient training. The input table stores the surface text of each
selected $n$-gram, together with the recipient-side lookup key used to index
the memory table. Following the notation in Section~\ref{sec:memory-grafting},
the selected memory keys are
\begin{equation}
	\mathcal{M}=\{\mathbf{m}_i\}_{i=1}^{M},
	\qquad
	\mathbf{m}_i=(x_{i,1},\ldots,x_{i,n_i}),
	\qquad n_i\in\{2,3,4\},
\end{equation}
where each $\mathbf{m}_i$ is the exact lookup key used by the recipient model.
The implementation stores a map from this key to the memory row index $i$.

\noindent To build the memory values, we run the grafting model offline on the surface
text of each $n$-gram. In the Qwen3.5-35B-A3B implementation, we tokenize the
$n$-gram text with the Qwen tokenizer, run the text backbone, and extract the
hidden state at the last non-padding token from selected grafting-model layers.
These vectors are written to disk and
then packed into frozen \texttt{nn.Embedding} tables in bfloat16 format. The
same construction also saves a map from the recipient-side $n$-gram id to the
row index of the stored vector. At runtime, the recipient only performs exact
lookup over $\mathcal{M}$, obtains the corresponding frozen vector, and uses
learned projections and gates to adapt it to the recipient hidden space.
Equivalently, for a selected grafting-model layer $r$, the stored value is
\begin{equation}
	\mathbf{W}_{\mathrm{mem}}[i]
	=F_G(\mathrm{text}(\mathbf{m}_i))_{\mathrm{last}}^{(r)}
	\in\mathbb{R}^{D_{\mathrm{mem}}},
\end{equation}
which is the implementation-level form of
$F_G(\mathbf{m}_i)_{\mathrm{last}}^{(r)}$ in Section~\ref{sec:memory-grafting}.
During recipient training and inference, the recipient queries this frozen
table with the current suffix context:
\begin{equation}
	(j_{b,t},m_{b,t})=\texttt{ExactLookup}(\mathbf{X}_{b,\leq t},\mathcal{M}),
	\qquad
	\mathbf{e}_{\mathrm{mem},b,t}=\mathbf{W}_{\mathrm{mem}}[j_{b,t}]
	\quad \text{if } m_{b,t}=1.
\end{equation}

\noindent This design avoids conflicts when the recipient model and the grafting model
use different tokenizers. Online retrieval is always performed over the
recipient-side key set $\mathcal{M}$, while the grafting-model tokenizer is
used only offline to encode the same surface text and produce the stored
latent value. Even if the grafting model splits the text differently, this
only changes the latent vector stored in $\mathbf{W}_{\mathrm{mem}}$; it does
not change the online lookup key. To make this explicit, let $\tau_G$ denote
the grafting-model tokenizer and let $s_i=\mathrm{text}(\mathbf{m}_i)$. Then
\begin{equation}
	\mathbf{W}_{\mathrm{mem}}[i]=F_G(\tau_G(s_i))_{\mathrm{last}}^{(r)},
\end{equation}
but online retrieval only queries $\mathcal{M}$ through $\mathbf{m}_i$. Thus,
$\tau_G(s_i)$ affects only the stored latent value, not the exact lookup key.

\subsection{Calculate resources and environment}

We use deepspeed as the training framework. For the 0.9B model, We conduct training on a cluster with 4 nodes and 32 A100 GPUs. For the 2.8B model, We conduct training on a cluster with 8 nodes and 64 A100 GPUs.

\clearpage

\subsection{Appendix: Details of the Grafting Gate}
\label{sec:further-analysis-grafting-gate-details}

We provide the detailed formulation of the grafting mechanisms used in the
ablation using the notation in Section~\ref{sec:method}. At recipient position
$(b,t)$ and branch $c$, let $\mathbf{H}_{b,t}^{c}$ denote the recipient hidden
state. For an exact matched $n$-gram order $n$, let
$\mathbf{e}_{\mathrm{mem},b,t}^{(n)}$ denote the corresponding frozen
grafting-model memory feature, and let $\mathcal{N}_{b,t}$ denote the set of
matched orders. The hash-based Engram fallback feature is
$\mathbf{e}_{\mathrm{hash},b,t}$. Given any candidate key-value pair
$(\mathbf{K}_{b,t}^{c},\mathbf{V}_{b,t})$, the Method gate is
\[
\mathbf{Q}_{b,t}^{c}=\mathrm{RMSNorm}_{q}(\mathbf{H}_{b,t}^{c}),
\qquad
\bar{\mathbf{K}}_{b,t}^{c}=\mathrm{RMSNorm}_{k}(\mathbf{K}_{b,t}^{c}),
\]
\[
\alpha_{b,t}^{c}=\sigma\left(
\frac{\left\langle\bar{\mathbf{K}}_{b,t}^{c},\mathbf{Q}_{b,t}^{c}\right\rangle}{\sqrt{D}}
\right),
\qquad
\mathbf{U}_{b,t}^{c}=\alpha_{b,t}^{c}\mathbf{V}_{b,t},
\]
\[
\Delta\mathbf{H}_{b,t}^{c}=\mathbf{U}_{b,t}^{c}
+\texttt{ShortConv}(\mathbf{U})_{b,t}^{c},
\qquad
\widetilde{\mathbf{H}}_{b,t}^{c}=\mathbf{H}_{b,t}^{c}+\Delta\mathbf{H}_{b,t}^{c} .
\]

\noindent \circled{1}  The attention-only grafting variant first aggregates all matched $n$-gram
embeddings and adds the resulting value without the query-key gate:
\[
\mathbf{U}_{\mathrm{attn},b,t}^{c} =\mathbf{e}_{\mathrm{attn},b,t}=
\texttt{Attn}\left(\mathbf{H}_{b,t}^{c},
\{\mathbf{U}_{\mathrm{attn},b,t}^{c} =\mathbf{e}_{\mathrm{mem},b,t}^{(n)}:n\in\mathcal{N}_{b,t}\}\right)
\]
\[
\Delta\mathbf{H}_{b,t}^{c}=\mathbf{U}_{\mathrm{attn},b,t}^{c}
+\texttt{ShortConv}(\mathbf{U}_{\mathrm{attn}})_{b,t}^{c} .
\]
\circled{2}  The attention-gated variant uses the same attention-based memory but modulates
it with the gate:
\[
\mathbf{K}_{\mathrm{attn},b,t}^{c}=\mathbf{W}_{k}^{\mathrm{mem},c}
\mathbf{e}_{\mathrm{attn},b,t},
\qquad
\mathbf{V}_{\mathrm{attn},b,t}=\mathbf{W}_{v}^{\mathrm{mem}}
\mathbf{e}_{\mathrm{attn},b,t},
\]
followed by the query-key gate above with
$(\mathbf{K}_{b,t}^{c},\mathbf{V}_{b,t})=(\mathbf{K}_{\mathrm{attn},b,t}^{c},
\mathbf{V}_{\mathrm{attn},b,t})$.

\noindent \circled{3}  The longest-match gated variant removes the attention aggregation and uses only
the longest exact match. For this ablation, missed positions do not use Engram
fallback; if no exact match is found, the lookup returns zero memory:
\[
(j_{b,t},m_{b,t})=\texttt{ExactLookup}(\mathbf{X}_{b,\leq t},\mathcal{M}),
\qquad
m_{b,t}=0 \quad \text{if no exact memory hit is found}.
\]
When $m_{b,t}=1$, let $n_{b,t}^*=\max\mathcal{N}_{b,t}$. The candidate memory
feature is then
\[
\mathbf{e}_{\mathrm{long},b,t}=
\begin{cases}
\mathbf{e}_{\mathrm{mem},b,t}^{(n_{b,t}^*)}, & m_{b,t}=1,\\
\mathbf{0}, & m_{b,t}=0,
\end{cases}
\]
and the projected key and value are
\[
\mathbf{K}_{\mathrm{long},b,t}^{c}=\mathbf{W}_{k}^{\mathrm{mem},c}
\mathbf{e}_{\mathrm{long},b,t},
\qquad
\mathbf{V}_{\mathrm{long},b,t}=\mathbf{W}_{v}^{\mathrm{mem}}
\mathbf{e}_{\mathrm{long},b,t},
\]
followed by the same query-key gate with
$(\mathbf{K}_{b,t}^{c},\mathbf{V}_{b,t})=(\mathbf{K}_{\mathrm{long},b,t}^{c},
\mathbf{V}_{\mathrm{long},b,t})$.

\noindent \circled{4}  The final mechanism adds Engram fallback to the gated longest-match mechanism:
\[
\mathbf{K}_{\mathrm{eng},b,t}^{c}=\mathbf{W}_{k}^{\mathrm{eng},c}
\mathbf{e}_{\mathrm{hash},b,t},
\qquad
\mathbf{V}_{\mathrm{eng},b,t}=\mathbf{W}_{v}^{\mathrm{eng}}
\mathbf{e}_{\mathrm{hash},b,t},
\]
\[
\mathbf{K}_{b,t}^{c}=m_{b,t}\mathbf{K}_{\mathrm{mem},b,t}^{c}
+(1-m_{b,t})\mathbf{K}_{\mathrm{eng},b,t}^{c},
\qquad
\mathbf{V}_{b,t}=m_{b,t}\mathbf{V}_{\mathrm{mem},b,t}
+(1-m_{b,t})\mathbf{V}_{\mathrm{eng},b,t},
\]
where $m_{b,t}$ is the exact-hit mask defined by
\texttt{ExactLookup}$(\mathbf{X}_{b,\leq t},\mathcal{M})$. The selected
$(\mathbf{K}_{b,t}^{c},\mathbf{V}_{b,t})$ is then passed to the Method gate.
This formulation makes the role of the gate explicit: it does not decide which
memory is retrieved, but controls how strongly the selected grafting signal is
written into the recipient hidden state through $\Delta\mathbf{H}$.
\clearpage

\section{Additional Experiment}
\subsection{Additional Analysis of Grafting-Model Memory Geometry}
\label{app:grafting-model-memory-geometry}

\begin{figure}[h!]
\centering
\includegraphics[width=0.95\linewidth]{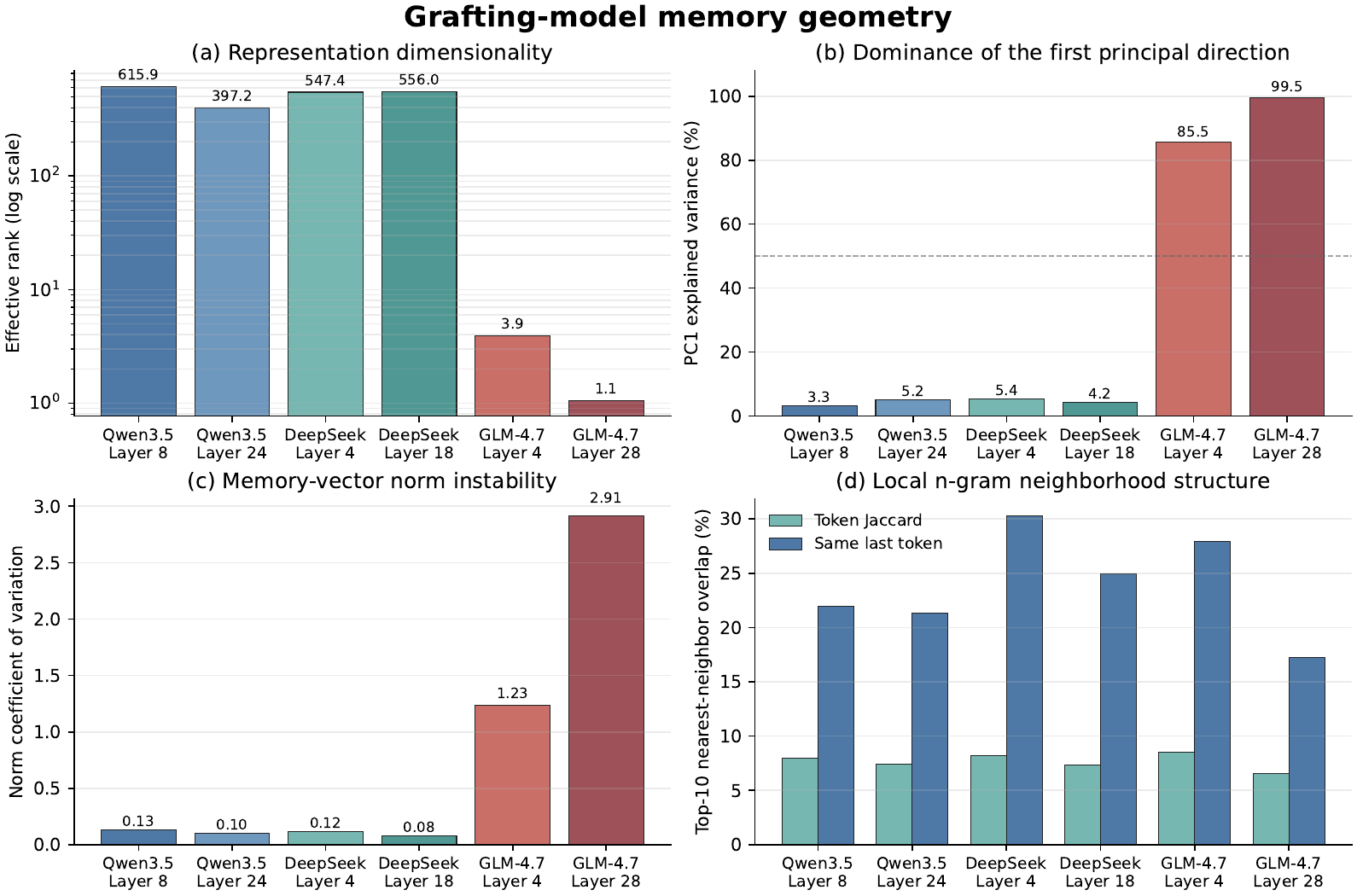}
\caption{Geometry of frozen grafting-model memory tables. We compare effective
rank, first-principal-component dominance, norm instability, and local
nearest-neighbor structure across memory tables constructed from Qwen3.5-35B-A3B,
DeepSeek-V2-Lite, and GLM-4.7-Flash.}
\label{fig:grafting_model_memory_geometry}
\end{figure}

\noindent Figure~\ref{fig:grafting_model_memory_geometry} compares frozen memory tables
constructed from the same $n$-gram key set but different grafting models:
Qwen3.5-35B-A3B Layers 8/24, DeepSeek-V2-Lite Layers 4/18, and GLM-4.7-Flash Layers
4/28. For each table, we sample 10,000 memory rows for norm statistics, 2,048
rows for effective rank and first-principal-component variance, and 1,024 rows
for cosine nearest-neighbor structure. All diagnostics are computed directly
on the stored memory vectors before recipient training.

\noindent Qwen3.5-35B-A3B and DeepSeek-V2-Lite produce high-dimensional and stable memory
geometry: their effective ranks remain in the hundreds, PC1 explains only a
small fraction of the variance, and norm variation is low. In contrast,
GLM-4.7-Flash is strongly anisotropic: Layer 4 already has low effective rank,
and Layer 28 is nearly dominated by a single direction with unstable norms.
Thus, many GLM $n$-grams retrieve nearly collinear vectors, making the memory
less discriminative. This suggests that Memory Grafting requires not only a
capable grafting model, but also structured local $n$-gram representations;
when this structure is weak, the improvement from grafting is limited.

\clearpage

\subsection{Loss Curves for Memory Grafting}
\label{sec:further-analysis-multilayer-loss}

\begin{figure*}[h!]
	\centering
	\includegraphics[width=1.0\linewidth]{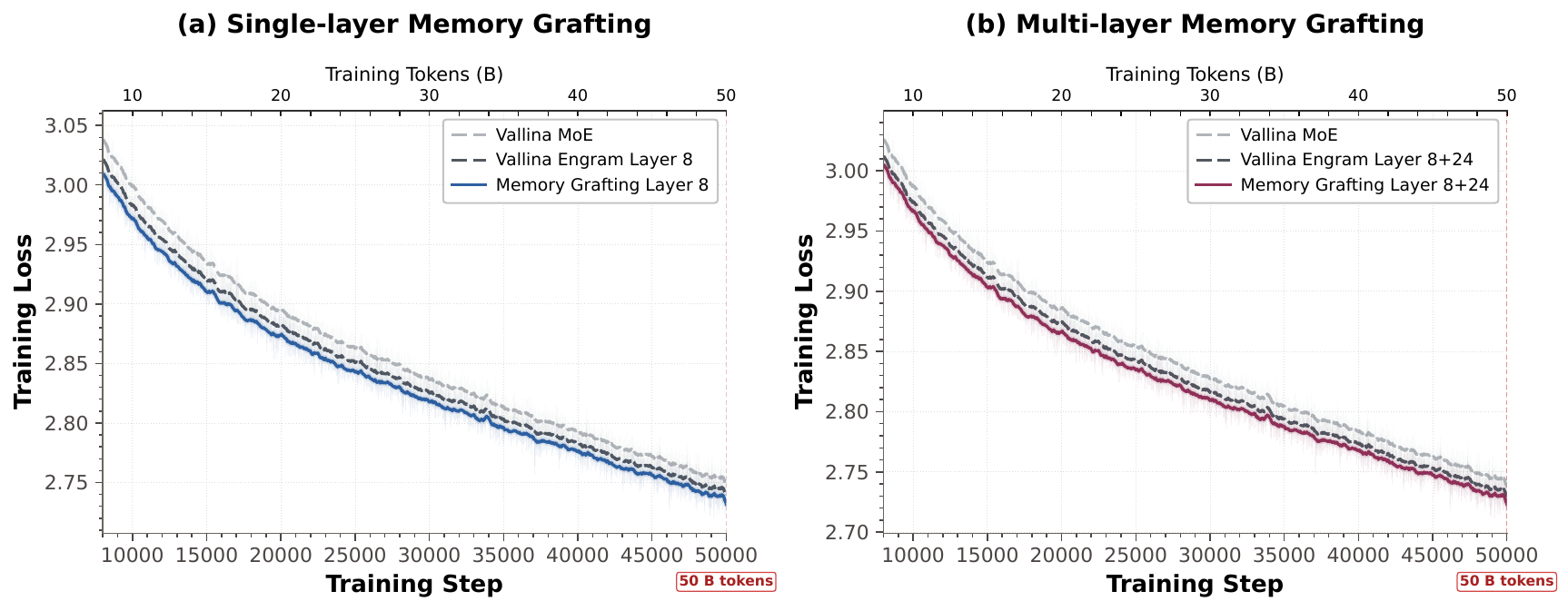}
	\caption{Training loss curves for memory graftings.}
	\label{fig:multilayer-loss}
\end{figure*}

\noindent We further report the training loss curves of single-layer and multi-layer
grafting settings in Figure~\ref{fig:multilayer-loss}. The results show that
Memory Grafting effectively reduces the training loss in both cases.

\clearpage
\subsection{More CKA Analysis Results of Memory Grafting}
\label{sec:more-cka-memory-grafting}

\begin{figure*}[h!]
	\centering
	\includegraphics[width=\linewidth]{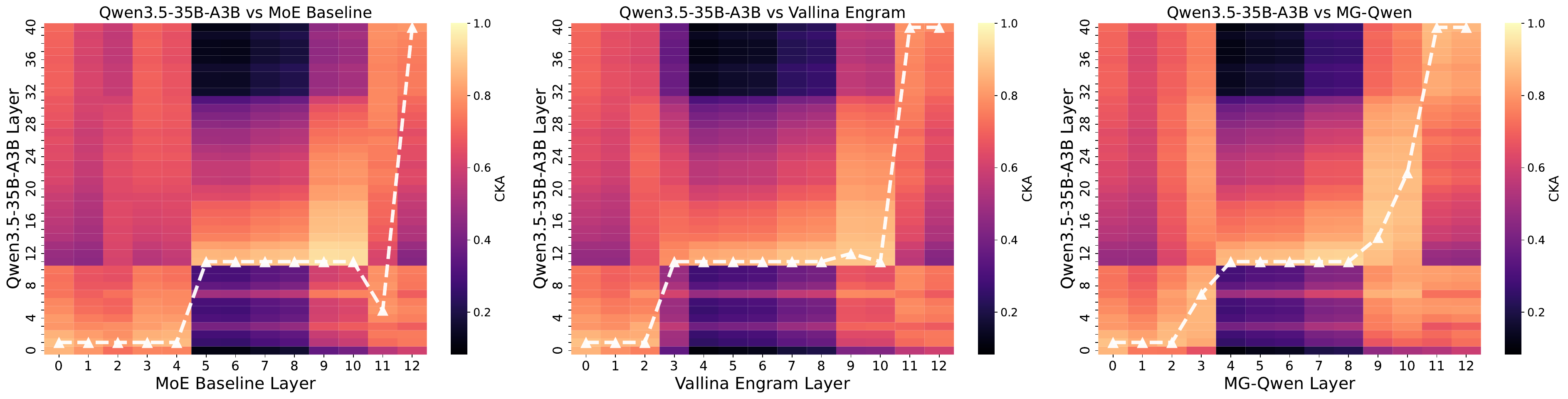}
	\caption{CKA comparison for single-layer Memory Grafting.}
	\label{fig:cka-memory-grafting-more-single}
\end{figure*}

\begin{figure*}[h!]
	\centering
	\includegraphics[width=\linewidth]{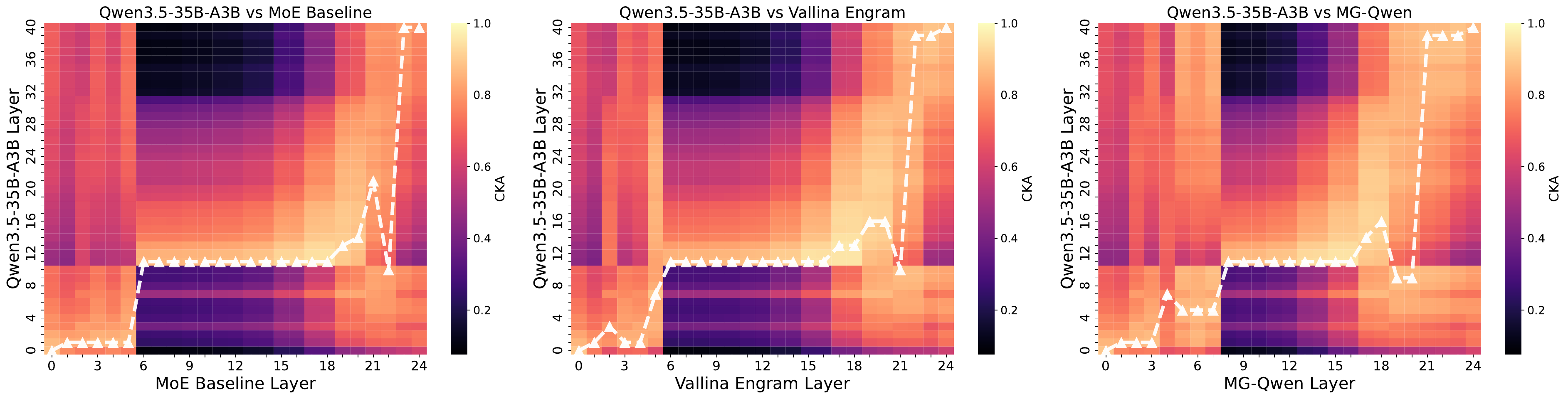}
	\caption{CKA comparison for the 2.8B model.}
	\label{fig:cka-memory-grafting-more-3b}
\end{figure*}

\noindent We provide additional CKA results in
Figures~\ref{fig:cka-memory-grafting-more-single} and
\ref{fig:cka-memory-grafting-more-3b}. Both the single-layer setting and the
2.8B Memory Grafting model show stronger alignment with deeper
Qwen3.5-35B-A3B representations than the corresponding baselines. This further
indicates that the grafted memory changes the recipient hidden states rather
than remaining an unused external table.

\clearpage
\subsection{Top-k N-gram Hit Rate}
\label{sec:further-analysis-topk-ngram-hit-rate}

\begin{figure*}[h!]
	\centering
	\includegraphics[width=\linewidth]{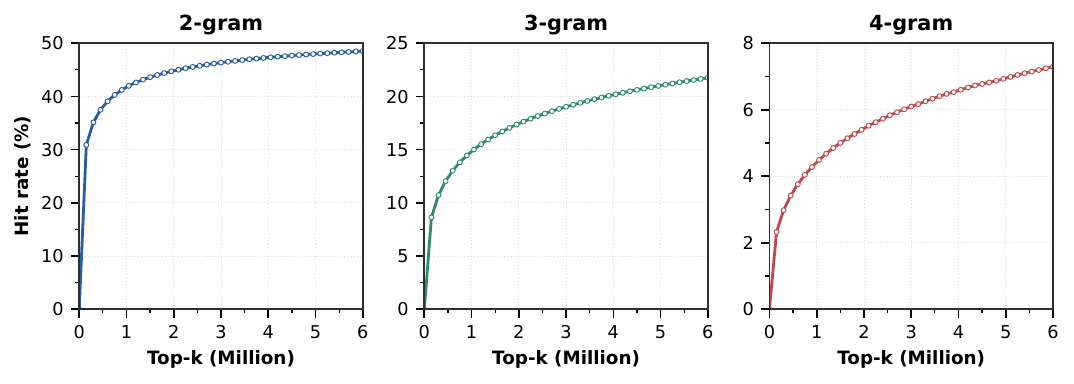}
	\caption{Hit rate of top-$k$ $n$-gram memory entries as the memory size
	increases.}
	\label{fig:topk-ngram-hit-rate}
\end{figure*}

\noindent We further analyze how the hit rate changes as more top-$k$ $n$-gram entries
are included in the grafting memory. As shown in
Figure~\ref{fig:topk-ngram-hit-rate}, the hit rate increases with the number of
stored $n$-grams, but the growth gradually slows down. This trend indicates
that frequent $n$-grams provide the largest coverage gain early on, while adding
more entries further improves coverage by capturing less frequent contexts.

\clearpage
\subsection{Single-Layer Memory Grafting Results}
\label{sec:further-analysis-single-layer-results}

\begin{table}[h]
\centering
\caption{Benchmark performance (\%) in the 0.8B-trainable, 50B-token setting. Only the first layer was grafted by the layer 6 of Qwen3.5-35B-A3B}
\label{tab:benchmark-single}
\setlength{\tabcolsep}{10pt}
\renewcommand{\arraystretch}{1.2}
\small
\resizebox{0.9\textwidth}{!}{
\begin{tabular}{@{}lc*{3}{c}}
\toprule
\textbf{Benchmark} \textit{\scriptsize(Metric)} & \textbf{\# Shots} & \textbf{MoE Baseline} & \textbf{Vanilla Engram} & \textbf{Memory Grafting} \\
\midrule
\# Trainable Params                          & & 0.92B          & 0.92B          & 0.92B \\
\# Activated \textit{\scriptsize(w/o token embed)} & & 0.29B          & 0.29B          & 0.29B \\
\# Trained Tokens                         & & 50B            & 50B            & 50B \\
\# Experts \textit{\scriptsize(shared + routed, top-$k$)} & & 1 + 56 (top-4) & 1 + 48 (top-4) & 1 + 46 (top-4) \\
\midrule
ARC-Challenge    & 0-shot  & 28.24 & 28.50 & \cellcolor{rowgray}\textbf{29.27} \\
ARC-Easy         & 0-shot  & 63.30 & 63.30 & \cellcolor{rowgray}\textbf{63.51} \\
BoolQ            & 0-shot  & 56.21 & 48.50 & \cellcolor{rowgray}\textbf{60.73} \\
Social IQA       & 0-shot  & 40.53 & 39.92 & \cellcolor{rowgray}\textbf{40.69} \\
RACE             & 0-shot  & 30.53 & 29.95 & \cellcolor{rowgray}\textbf{32.06} \\
LAMBADA          & 5-shot  & 29.50 & 29.11 & \cellcolor{rowgray}\textbf{30.37} \\
WinoGrande       & 5-shot  & 51.78 & 52.09 & \cellcolor{rowgray}\textbf{52.72} \\
PIQA             & 5-shot  & 70.24 & 71.22 & \cellcolor{rowgray}\textbf{71.49} \\
HellaSwag        & 10-shot & 38.03 & 38.25 & \cellcolor{rowgray}\textbf{38.40} \\
\midrule
\textbf{Average} & & 45.37 & 44.54 & \cellcolor{rowgray}\textbf{46.58} \\
\bottomrule
\end{tabular}
}
\end{table}

\noindent We also evaluate Memory Grafting under the single-layer grafting setting. With
the same model size and activated computation, Memory Grafting achieves the
best average performance among the compared models, improving over both the MoE
baseline and vanilla Engram. The gains are consistent across the evaluated
benchmarks, showing that even a single grafting layer can effectively transfer
useful $n$-gram memory into the recipient model.

\clearpage
\subsection{Probe Analysis}
\label{sec:further-analysis-memory-probe}

\begin{figure*}[h!]
	\centering
	\includegraphics[width=\linewidth]{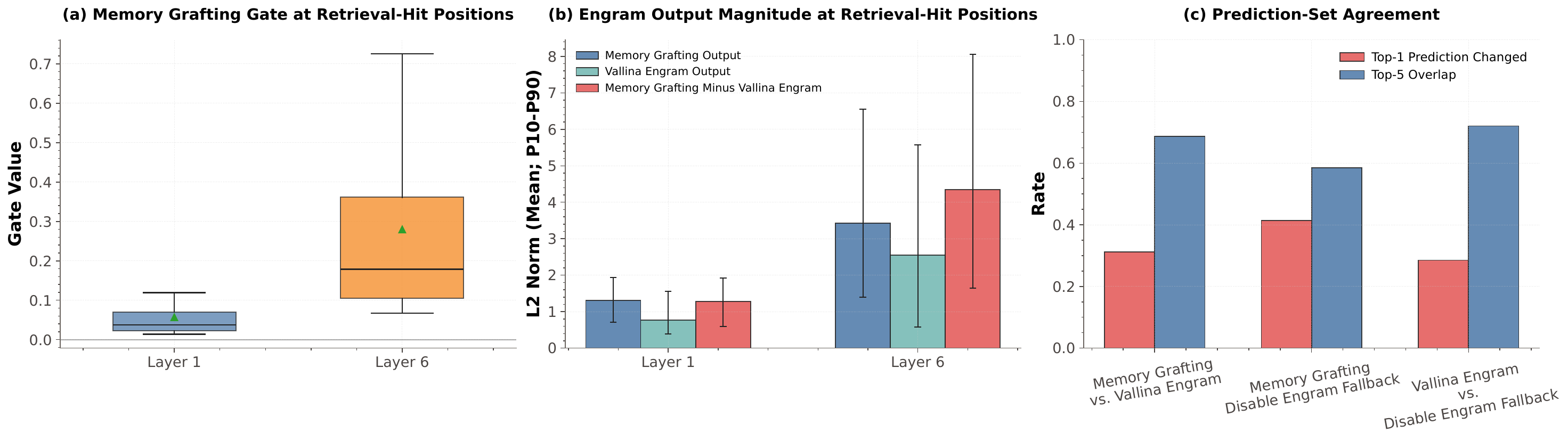}
	\caption{Overview of gate values, output magnitudes, and prediction-set
	agreement at retrieval-hit positions.}
	\label{fig:memory-probe-overview}
\end{figure*}

\begin{figure*}[h!]
	\centering
	\includegraphics[width=\linewidth]{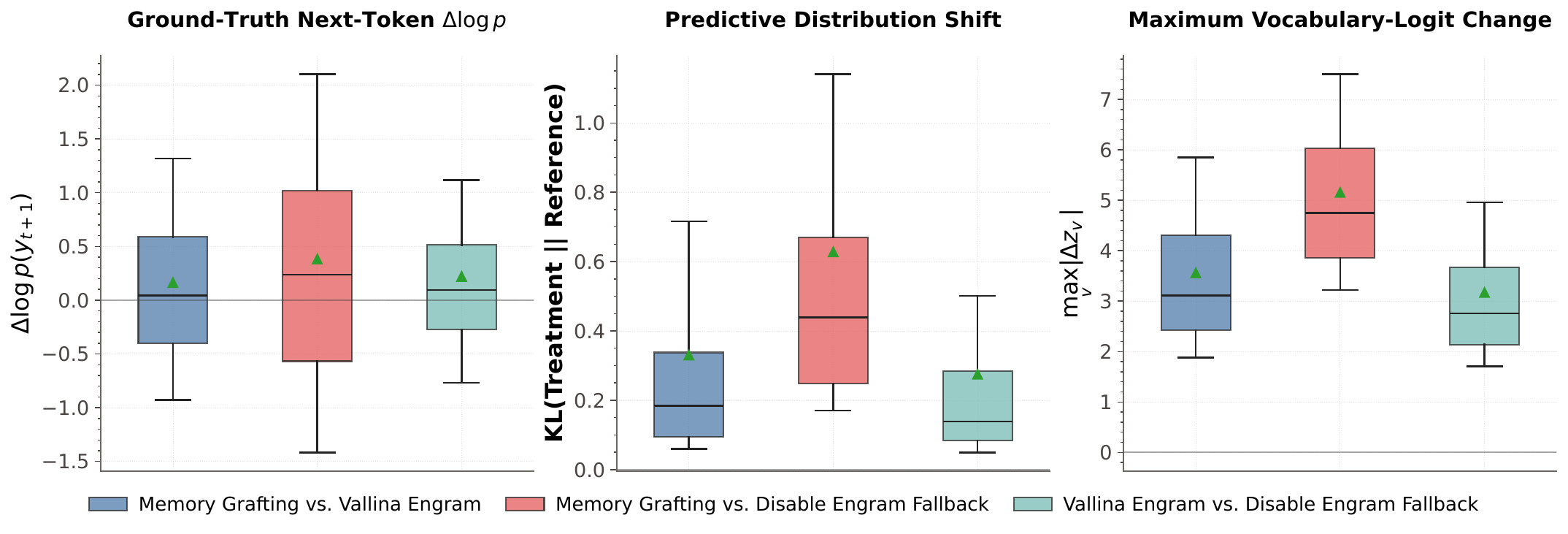}
	\caption{Logit-level impact of grafted memory at retrieval-hit positions.}
	\label{fig:memory-probe-logit-impact}
\end{figure*}

\noindent We separately probe the behavior of the grafted memory at retrieval-hit
positions, where an exact $n$-gram match is found in the external memory.
Figure~\ref{fig:memory-probe-overview} shows that the gate is non-zero and the
memory output has a clear magnitude at these positions, meaning that the
retrieved memory is actually written into the recipient hidden states.
Figure~\ref{fig:memory-probe-logit-impact} further shows that this hidden-state
change propagates to the output distribution: the next-token logits, target
token log-probability, and prediction set all change after grafting. Therefore,
the grafted memory is not only retrieved, but also participates in the model's
computation and directly affects its predictions.

\clearpage
\section{Limitations \& Discussion}
\subsection{Limitations}
We do not include a full comparison with knowledge distillation. A from-scratch
distillation baseline at the same pre-training scale would require running a
large teacher model over the full training corpus and repeating this process
under comparable recipient settings, which would require substantial
additional computation and is difficult to complete under our compute budget.
Moreover, forcing distillation into the same compute budget would also be
difficult to interpret, because the result may reflect insufficient
distillation tuning or an under-resourced distillation setup rather than a
clean comparison of the transfer mechanism. Such a comparison would therefore
not provide a clean evaluation of the effectiveness of Memory Grafting.

\noindent To make this gap concrete, For example, distilling a 1B-A0.2B recipient from a 30B-A3B teacher requires
one extra teacher forward pass for every training token. Using standard
transformer compute accounting\citep{kaplan2020scaling,hoffmann2022training}, the recipient's own training cost is roughly
$6\times0.2$B activated computation per token, while the teacher forward adds
$2\times3$B. Thus, online distillation would add about $5\times$ the
recipient's own training compute, making the total cost about $6\times$ solo
recipient pre-training.

\noindent Therefore, our experiments focus on comparisons under nearly matched recipient
architectures, token budgets, and activated computation. Under these controlled
conditions, Memory Grafting serves as a knowledge-transfer method independent
of knowledge distillation, showing that reusable latent memory can improve
pre-training without running a teacher model online.

\subsection{Future Work}

\textbf{Scaling the grafted memory.}
The capacity sweep in Figure~\ref{fig:memory-capacity-scaling} shows that loss keeps
decreasing from $30$k to $3$M entries, and the hit-rate curves are still
rising. Pushing the bank to larger $n$-gram counts and higher orders, and
sweeping it together with recipient size and token budget, is a direct way to
extend Memory Grafting to frontier scales.

\noindent\textbf{Memory Grafting for post-training and continual learning.}
Beyond from-scratch pre-training, the same construction can serve as a
post-training and continual-learning strategy~\cite{ran2025users,ran2025understanding,chen2024exploring,wei2025modeling,wei2025unifying,xiong2024multi,cheng2025whoever,guan2025enhancinglogitsdistillationplugplay,xiong2025hs,wei2025learning}..Because the bank is built
offline at modest cost (about $3$ A100-hours for a $3$M-entry table in our
setting) and the recipient consumes it only through frozen retrieval plus
lightweight projections and gates, a freshly built bank can be attached to an
already-trained recipient to inject domain- or task-specific knowledge during
SFT, instruction tuning, or domain adaptation, without re-running large-scale
pre-training. In a continual-learning setting, the bank can be incrementally
extended or rebuilt as new corpora and stronger grafting models become
available, letting the recipient absorb new knowledge through its external
memory rather than through parameter updates that risk catastrophic
forgetting.

\noindent\textbf{Grafting mechanism and recipient-side structure.}
The current interface---longest-match retrieval, linear projections, a
query-key gate, a short convolution, and a residual update at two layers---is
already effective, and the ablation in Figure~\ref{fig:gate_ablation} shows
that each component matters. Richer aggregation over multiple matched
$n$-grams, gates conditioned on retrieval properties, alternatives to the
additive residual update, and adaptive recipient-layer selection are all
plausible next steps on top of this design.

\section{LLM Usage}
This study utilizes Large Language Models (LLMs) to refine content, adjust formatting, construct
tables, and provide writing suggestions for specific chapters.



\end{document}